\documentclass{article}

\usepackage{arxiv}

\usepackage[T1]{fontenc}    
\usepackage{url}            
\usepackage{booktabs}       
\usepackage{amsfonts}       
\usepackage{nicefrac}       
\usepackage{microtype}      
\usepackage{lipsum}
\usepackage{graphicx}
\usepackage{subfig}
\RequirePackage{amsmath,amsfonts,amssymb}
\RequirePackage{mathptmx}

\RequirePackage{xcolor}
\RequirePackage{authblk}
\RequirePackage[latin1,utf8]{inputenc}
\RequirePackage[english]{babel}
\RequirePackage{lmodern}
\RequirePackage{siunitx}
\RequirePackage{textgreek}

\RequirePackage{gensymb}
\RequirePackage{textcomp}
\RequirePackage[version=4]{mhchem}
\RequirePackage[scaled]{helvet}
\RequirePackage[T1]{fontenc}
\RequirePackage{lettrine} 
\RequirePackage[rightcaption]{sidecap} 
\RequirePackage[misc]{ifsym} 
\RequirePackage{bbding} 

\DeclareUnicodeCharacter{00B0}{\degree}

\RequirePackage[colorlinks=true, allcolors=blue]{hyperref}

\RequirePackage{cleveref}

\title{Evaluation of Deep Convolutional Generative Adversarial Networks for data augmentation of chest X-ray images}

\author{
  Sagar Kora Venu \\
  Department of Analytics and Data Science\\
  Harrisburg University of Science and Technology\\
  Harrisburg, PA 17101 \\
  \texttt{SKora@my.harrisburgu.edu}
}

\begin{document}
\maketitle

\begin{abstract}
Medical image datasets are usually imbalanced, due to high costs of obtaining the data and time-consuming annotations. Training deep neural network model on such datasets to accurately classify the medical condition does not yield desired results and often over-fits the data on majority class samples. In order to address this issue, data augmentation is often performed on training data by position augmentation techniques such as scaling, cropping, flipping, padding, rotation, translation, affine transformation, and color augmentation techniques such as brightness , contrast, saturation, and hue to increase the dataset sizes. These augmentation techniques are not guaranteed to be advantageous in domains with limited data, especially medical image data, and could lead to further overfitting. In this work, we performed data augmentation on Chest X-rays dataset through generative modeling (deep convolutional generative adversarial network) which creates artificial instances retaining similar characteristics to the original data and evaluation of the model resulted in Fréchet Distance of Inception (FID) of 1.289.
\end{abstract}


\section*{Introduction}
\begin{figure}[htbp] 
\centering
\includegraphics[scale=0.5]{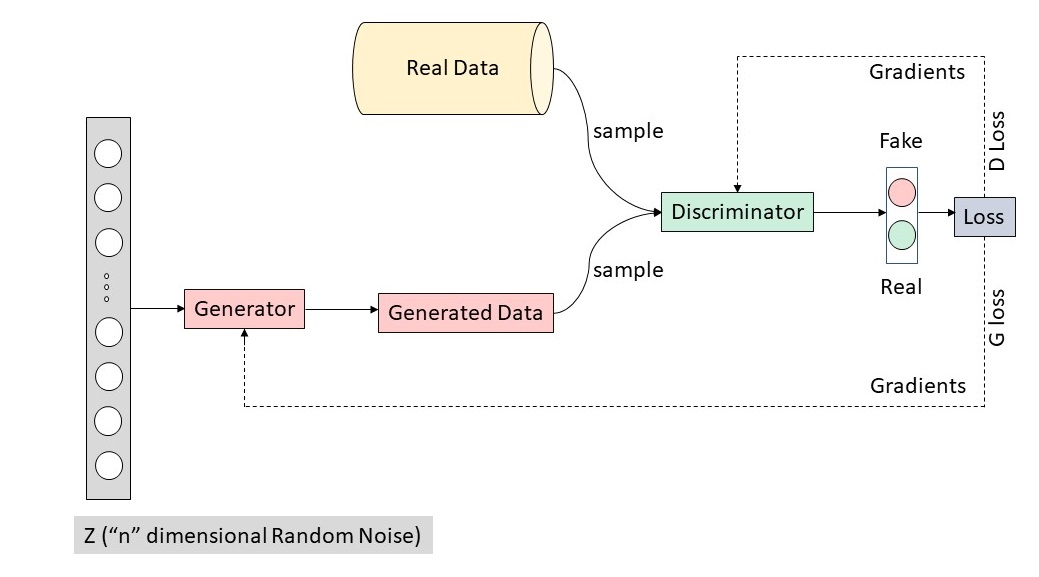}
\caption{Generative Adversarial Network Architecture}
\label{fig:GAN}
\end{figure}

Data sets for medical imaging are limited in size due to privacy issues and getting annotation of medical images is expensive and time-consuming, which often leads to having only small amounts of labeled medical imaging data to use for image classification tasks. Deep learning techniques need a huge volume of data to train effective models for tasks such as image recognition/ classification. Data augmentation is a technique commonly used in deep learning to expand data and prevent over-fitting in such data-limited situations. In this work, we investigate the use of Deep Convolutional Generative Adversarial Networks for generating chest X-ray images to augment the original dataset\footnote{All code, hyper-parameters may be found at https://github.com/sagarkora/DCGAN-ChestXray}. Generative Adversarial Networks (GAN’s) were introduced by Ian Goodfellow and his colleagues in 2014 \cite{goodfellow2014generative}. GAN’s utilize two neural networks, a generator which takes random noise as input to create samples (data) as realistic as possible to the original dataset and a discriminator to distinguish between data that is real (original data) vs fake (generated data) as shown in Figure~\ref{fig:GAN}.

The DCGAN proposed by \cite{radford2015unsupervised} is direct extension of the original GAN \cite{goodfellow2014generative}, except that the discriminator and generator explicitly uses convolutional and convolutional-transpose layers, respectively.

\section*{Materials and Methods}

\begin{figure}[htbp]
\centering
\includegraphics[width=\textwidth]{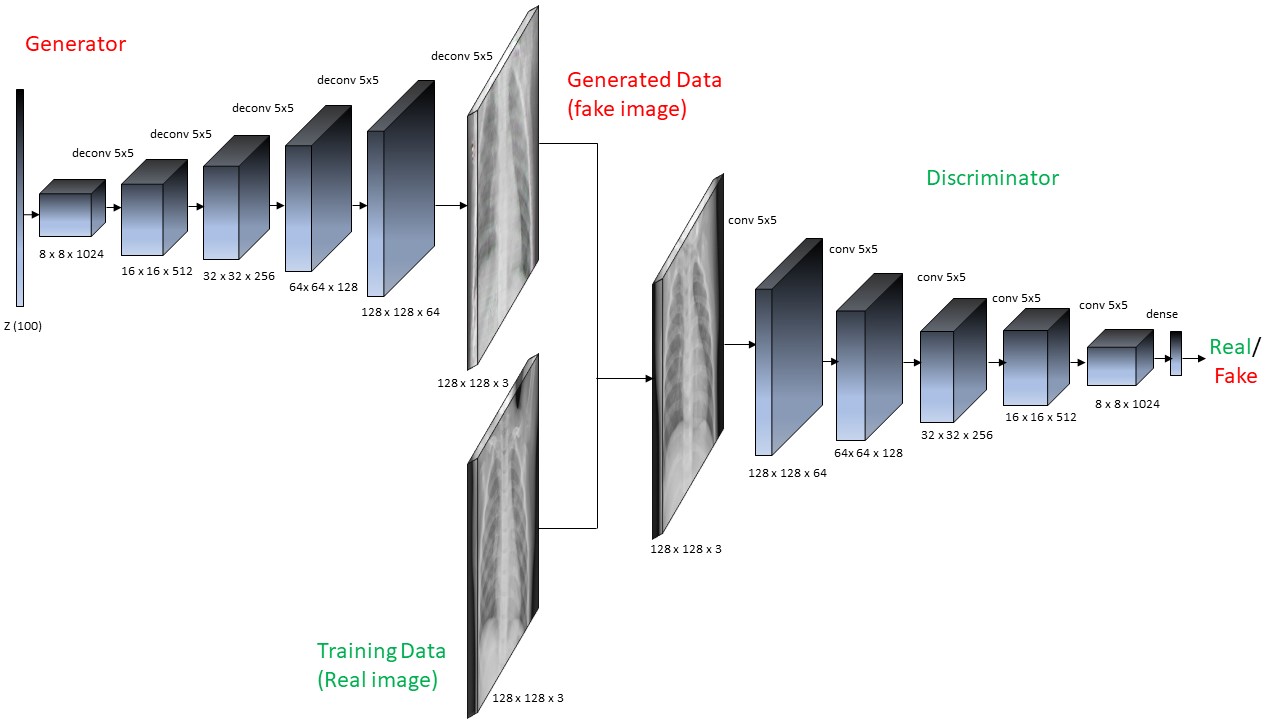}
\caption{Deep Convolutional Generative Adversarial Network Architecture}
\label{fig:DCGAN}
\end{figure}

We used X-ray images data obtained by \cite{kermany2018identifying} in the experiment. The dataset was already organized into three folders (train, test, val) and each folder contained sub-folders for each image category (Normal/ Pneumonia) with 5216 X-ray images in the train folder (1341 images are labeled with Normal and 3875 images labeled with Pneumonia), 16 X-ray images in val folder and 624 X-ray images in test folder. Its obvious that the data in the train folder is imbalanced and training a neural network to classify the data among two categories will over-fits the data . So, in this experiment, we augment the Normal X-ray images by Deep Convolutional Generative Adversarial Networks with the architecture as shown in Figure~\ref{fig:DCGAN}.
The images were resized to 128x128 pixels due to GPU memory constraints and then the images are scaled to a [-1,1] pixel value range to match the output of the generator as it uses Tanh activation function. In this architecture, a 100x1 noise vector is fed as an input to the generator. There are then four Convolutional layers with 2D-upsampling layers applied with Leaky ReLu activation function interlaced in between to scale to the appropriate 128x128 image size. The discriminator network is a similar network with four convolutional layers and a stride of 2 with leaky ReLU as an activation function except for the final node which is a sigmoid activation function to output if the image is real (original data) or fake (generated data). 

\subsection*{GAN Objective Function}
Let $\boldsymbol{z}$ be the latent space vector sampled from a standard normal distribution, $G(\boldsymbol{z})$ represents the generator function which maps the latent vector $\boldsymbol{z}$ to data-space, $\boldsymbol{x}$ be the data representing an image, and $D(\boldsymbol{x})$ is the discriminator network which outputs the probability that $\boldsymbol{x}$ came from the training data (real) rather than the generated data (fake) from the generator distribution $p_{\boldsymbol{z}}(\boldsymbol{z})$. 
The goal of $\boldsymbol{G}$ is to estimate the distribution that the training data came from $p_{\text{data}}$ so it can generate fake samples from that estimated distribution $p_{\text{g}}$ \cite{goodfellow2014generative}. 
\par
The learning process of the GANs is to train a discriminator and a generator simultaneously, which is a mini-max game between discriminator and generator where, the discriminator tries to maximize the loss function, i.e., $\boldsymbol{D}$ tries to maximize the probability that it correctly classifies real images and fake images $(\log D(\boldsymbol{x}))$ and the generator tries to minimize the loss function, i.e., $\boldsymbol{G}$ tries to minimize the probability that $\boldsymbol{D}$ will predict its outputs are fake $(\log(1-D(G(\boldsymbol{z}))))$  as shown in the equation (\ref{GANLossfn}). 

\begin{equation} 
\min_{G}\ \max_{D}V_{\mathrm{GAN}}(D,\ G)=\mathbb{E}_{\boldsymbol{x}\sim p_{\text{data}}(\boldsymbol{x})}[\log D(\boldsymbol{x})] +\mathbb{E}_{\boldsymbol{z}\sim p_{\boldsymbol{z}}(\boldsymbol{z})}[\log(1-D(G(\boldsymbol{z})))]. 
\label{GANLossfn} 
\end{equation}

Theoretically, the solution to this mini-max game is where $p_{\text{g}}$ = $p_{\text{data}}$, such that the discriminator guesses randomly if the inputs are real images (training data) or fake images (generated data). However, GANs' theory of convergence is still being actively studied, and in fact models have not always been trained to this extent.


\section*{Results}

The DCGAN was trained for 500 epochs and in just around 50 epochs, the DCGAN was able to generate images that resembled the chest X-ray images and then the quality of generated images further improved over 500 epochs. 
\begin{figure}[htbp]
\centering
\includegraphics[width=\textwidth]{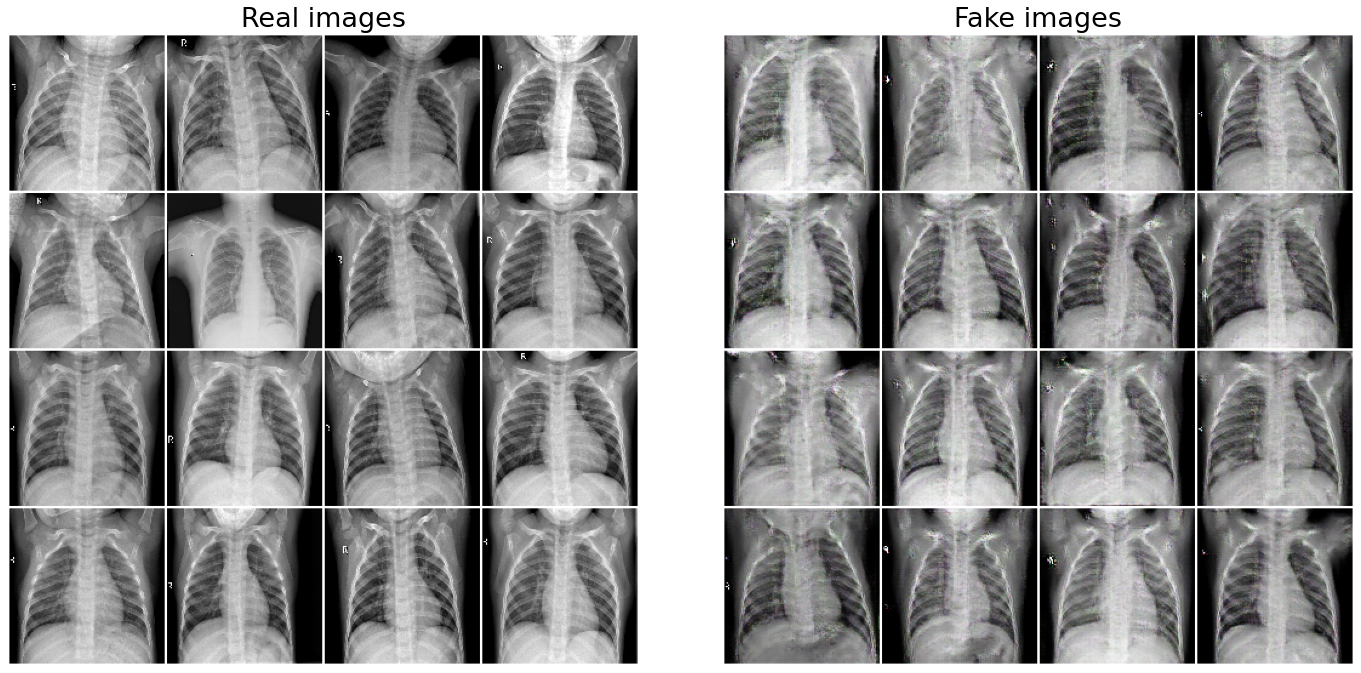}
\caption{Images from Original dataset (Real) and Images generated by the generator of DCGAN (fake)}
\label{fig:Result}
\end{figure}
For comparison, we show a grid of Real images (original data) and the fake images (generated images) in the Figure~\ref{fig:Result}.

\begin{figure}[htbp]
    \centering
    \subfloat[Loss]{{\includegraphics[width=7cm]{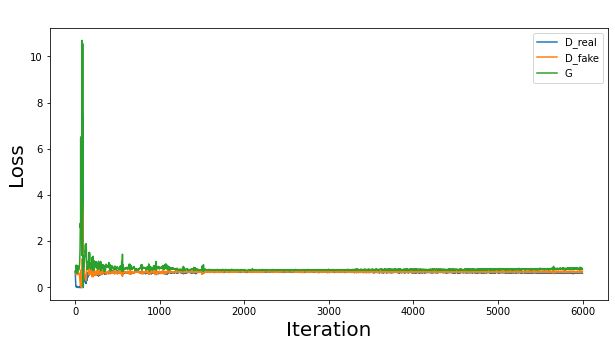} }}%
    \qquad
    \subfloat[Accuracy]{{\includegraphics[width=7cm]{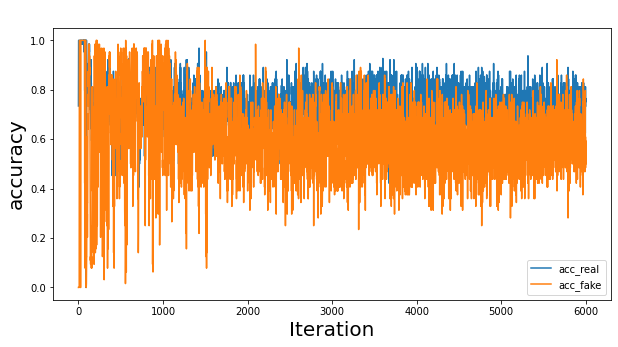} }}%
    \caption{DCGAN Loss and Accuracy during training}%
    \label{fig:DCGANlossacc}%
\end{figure}

The loss and accuracy of the generator and discriminator during training is shown in Figure~\ref{fig:DCGANlossacc}.

\subsection*{Evaluation}

Deep learning models are usually trained with a loss function until neural network convergence. Since GAN's are trained with two neural networks simultaneously to reach a Nash Equilibrium, there is no objective loss function to train GAN generator models to objectively access the training progress and quality of the model from the loss of the discriminator network and/ or the loss of the generator network \cite{salimans2016improved}. In general, to access the quality of the generated images based on the performance of the GAN models, two techniques have been developed: 1. Quantitative Measures such as Average Log-likelihood, Inception Score (IS) \cite{salimans2016improved}, Fréchet Inception Distance (FID) \cite{heusel2017gans}, Maximum Mean Discrepancy (MMD) \cite{ASENS_1953_3_70_3_267_0} etc. and 2. Qualitative Measures such as Nearest Neighbours, Rating and Preference Judgment, Evaluating Mode Drop and Mode Collapse \cite{srivastava2017veegan} etc. Initiation Score (IS) and Fréchet Distance of Inception (FID) are two of the GAN evaluation measures that are widely accepted \cite{borji2018pros}. In this work, we evaluate the DCGAN model using Fréchet Distance of Inception (FID) measure.

\subsubsection*{Fréchet Distance of Inception (FID)}

Fréchet Distance of Inception (FID) score is a measure used to evaluate the performance of the Generative Adversarial Network based on the quality of generated images which captures the similarity of the generated images to the real images proposed by \cite{heusel2017gans} as an improvement to the Inception Score (IS) \cite{salimans2016improved}. FID score is calculated using the statistics of generated images to real images using the Fréchet distance also known as Wasserstein-2 distance between the two multivariate Gaussian's as shown in the equation (\ref{FID})

\begin{equation}
{d_{FID}}(x,g) = {\left\| {{\mu _x} - {\mu _g}} \right\|^2} + Tr\left[ {{\Sigma _x} + {\Sigma _g} - 2{{\left( {{\Sigma _x}{\Sigma _g}} \right)}^{\frac{1}{2}}}} \right]
\label{FID} 
\end{equation}

where ${\mu _x}$ and ${\mu _g}$ are the feature-wise mean of real and generated images respectively, ${\Sigma _x}$ and ${\Sigma _g}$ are the covariance matrix of real and generated images repectively, $Tr$ is the trace whichis the sum of the elements along the main diagonal of the square matrix, $X_{x} \sim \mathcal{N}(\mu _x, \Sigma _x)$ and $X_{g} \sim \mathcal{N}(\mu _g, \Sigma _g)$ are the 2048-dimensional activation's of the Inception-V3 pool3 layer for real images and generated images respectively. 
\par
To calculate the Gaussian statistics (mean and covariance), the number of samples (real images and generated images respectively) should be greater than the dimension of the coding layer i.e., the samples should be greater than 2048 for the Inception-V3 pool 3 layer, otherwise the covariance is not full rank resulting in complex numbers and NAN's.  Since, we had very limited samples (less than 2048) in our training dataset, we could not take advantage of the Inception-V3 pool3 layer, so we used the previous layer which is a Pre-aux classifier that is a 768-dimensional feature. We then calculated the Fréchet Distance of Inception (FID) score using \cite{pytorchf86:online} and the model achieved a FID score of 1.289 (lower scores correspond to better GAN performance).


\section*{Conclusions}

The contributions of Generative Adversarial Networks to the field of Medical imaging are highly appreciated, especially where there is limited access to the medical imaging data and and the high costs of obtaining  the labeled data. In this study, we applied deep convolutional generative adversarial networks (DCGAN) to generate artificial instances of chest X-ray images of the under-represented class in the dataset that resemble the chest X-ray images from the original dataset and evaluated the model using Fréchet Distance of Inception (FID) achieving a score of 1.289.


\section*{Forthcoming Research}

\begin{itemize}
\item To test the visual quality of the generated X-ray images, we intend to supply the generated images to a clinician to label the images as either real or fake (generated).
\item Develop a deep convolutional neural network to improve the accuracy in classifying the medical condition by utilizing the generated images alongside real training images.
\end{itemize}

\subsubsection*{Acknowledgements}

We thank \cite{kermany2018identifying} for making the datasets publicly accessible and we also thank Harrisburg University of Science and Technology for their support.

\bibliographystyle{unsrt}  
\bibliography{references} 
\end{document}